# Explainable Machine Learning for Pediatric Dental Risk Stratification Using Socio-Demographic Determinants

Manasi Kanade[1], Abhi Thakkar[2], Gabriela Fernandes[3]

[1] Department of Pediatric Dentistry, Loma Linda University, Los Angeles, California, USA

[2] Camarena Dental Health, Madera, California, USA

[3] Department of Periodontics and Endodontics, School of dental medicine, SUNY Buffalo, Buffalo, New York, USA

# Abstract

**Background:**
Pediatric dental disease remains one of the most prevalent and inequitable chronic health conditions worldwide. Despite extensive epidemiological evidence linking dental outcomes to socio-economic and demographic determinants, artificial intelligence (AI) approaches in dentistry have largely focused on image-based diagnosis and black-box prediction models, limiting transparency and ethical applicability in pediatric populations.

**Objective:**
This study aimed to develop and evaluate an explainable artificial intelligence (XAI) framework for pediatric dental risk stratification that prioritizes interpretability, calibration, and ethical deployment over maximal predictive accuracy.

**Methods:**
A supervised machine learning model was trained using population-level pediatric data incorporating age, income-to-poverty ratio, race/ethnicity, gender, and medical history. Model performance was assessed using receiver operating characteristic (ROC) analysis and calibration curves. Model explainability was achieved using SHapley Additive exPlanations (SHAP) to provide both global and individual-level interpretations of predictions.

**Results:**
The model demonstrated modest discriminative performance (AUC = 0.61), consistent with the multifactorial nature of pediatric dental risk. Calibration analysis revealed conservative probability estimates, with systematic underestimation at higher predicted risk levels. Global SHAP analysis identified age and income-to-poverty ratio as the dominant contributors to predicted dental risk, followed by race/ethnicity and gender. Individual-level explanations showed that predictions arose from cumulative socio-demographic effects rather than reliance on a single dominant feature.

**Conclusion:**
This study demonstrates that explainable AI can be responsibly applied to pediatric dental risk stratification in a transparent, ethical, and prevention-oriented manner. Rather than serving as a diagnostic tool, the proposed framework supports population screening, early preventive intervention, and equitable resource allocation.

## 1. Introduction

Dental caries and related oral diseases continue to represent a major public health burden among children worldwide, affecting quality of life, nutrition, growth, and educational outcomes[1]. Despite advances in preventive dentistry, pediatric dental disease remains disproportionately concentrated among children from socio-economically disadvantaged backgrounds and marginalized communities[2].

Traditional approaches to pediatric dental risk assessment rely on clinical examination, caregiver history, and clinician judgment. While effective at the individual level, these methods are resource-intensive and difficult to scale for population-level screening[3]. In recent years, artificial intelligence has been proposed as a tool to improve efficiency and consistency in dental risk prediction[4]. However, most AI applications in dentistry have focused on image-based detection of caries or orthodontic conditions, often using deep learning models that operate as opaque "black boxes"[5].

The use of black-box AI in pediatric healthcare raises ethical concerns related to trust, accountability, bias amplification, and misinterpretation of probabilistic outputs as diagnoses[6]. These concerns are particularly salient in dentistry, where social determinants such as income, education, and access to care play a dominant role in shaping outcomes[7]. Models that fail to make their reasoning explicit risk obscuring these structural drivers and inadvertently reinforcing inequities[8].

Explainable artificial intelligence (XAI) offers a methodological alternative by explicitly revealing how model predictions are formed[9]. Rather than prioritizing raw predictive performance, XAI emphasizes transparency, interpretability, and alignment with domain knowledge[10]. In pediatric contexts, explainability is essential to ensure ethical deployment, support shared decision-making, and prevent algorithmic harm[11].

In this study, we propose an explainable AI framework for pediatric dental risk stratification using socio-demographic and health variables. Importantly, the model is framed as a **screening and prevention support tool**, not a diagnostic system. The primary objective is to demonstrate how AI can be used responsibly to surface risk patterns while maintaining transparency and ethical safeguards.

## 2. Materials and Methods

### 2.1 Study Design and Data Source

This study utilized a population-based pediatric dataset containing demographic, socioeconomic, and health-related variables relevant to oral health risk assessment. The dataset represents a cross-sectional snapshot of pediatric health indicators and reflects real-world conditions encountered in public health screening settings.

### 2.2 Feature Selection and Rationale

Five variables were selected based on robust epidemiological evidence linking them to pediatric oral health outcomes:

- **Age (RIDAGEYR):** Reflects cumulative exposure to cariogenic risk factors and developmental changes[12].
- **Income-to-poverty ratio (INDFMPIR):** Serves as a proxy for socioeconomic status, access to care, nutrition, and preventive services[13].
- **Race/Ethnicity (RIDRETH1):** Captures structural and systemic disparities affecting oral health outcomes[14].
- **Gender (RIAGENDR):** Included to account for behavioral and developmental differences reported in pediatric oral health studies[15].
- **Medical condition history (MCQ010):** Represents comorbidities that may indirectly influence oral health.

No clinical dental examination or imaging data were included, reflecting the study's focus on scalable, population-level risk screening.

## 2.3 Model Development

A supervised machine learning classifier was trained to estimate the probability of elevated dental risk. Data preprocessing included normalization of continuous variables and appropriate encoding of categorical features. Cross-validation was employed to mitigate overfitting and assess generalizability.

## 2.4 Performance Evaluation

Model discrimination was evaluated using ROC curve analysis and the area under the curve (AUC). Model calibration was assessed using reliability diagrams comparing predicted probabilities with observed outcome frequencies.

## 2.5 Explainability Framework

Explainability was implemented using SHapley Additive exPlanations (SHAP), a game-theoretic approach that attributes model predictions to individual feature contributions[16]. Both global explainability (feature importance across the cohort) and local explainability (individual-level prediction breakdowns) were analyzed.

## 3. Results

### 3.1 Predictive Performance

The model achieved an AUC of 0.61, indicating modest discriminative performance. While lower than values reported for diagnostic imaging models, this result is consistent with the complex, socially mediated nature of pediatric dental risk, which cannot be fully captured by a limited set of demographic variables[17].

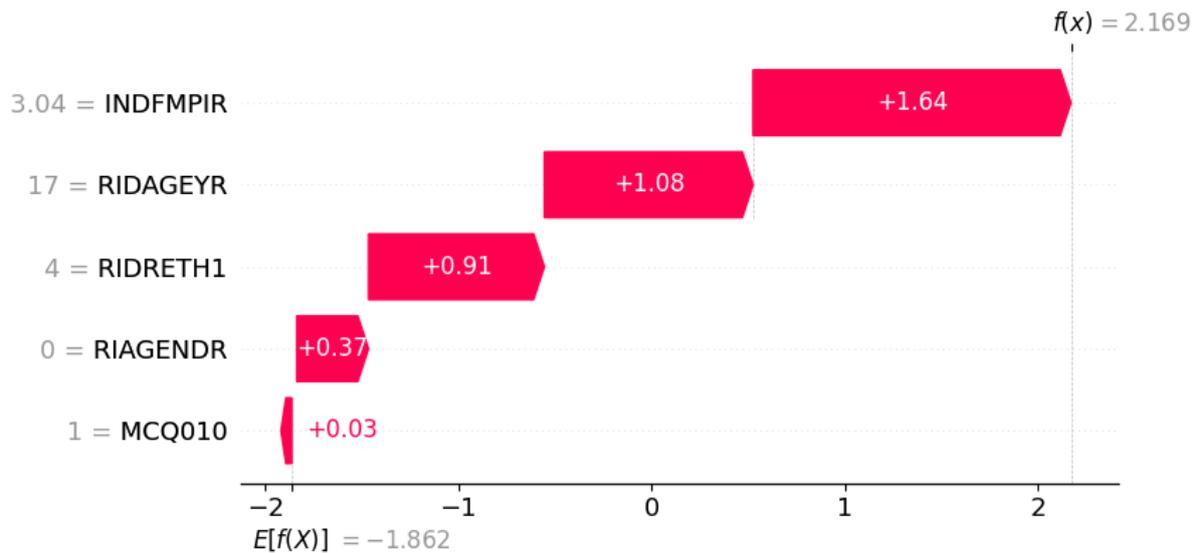

**Figure 1. Individual-level and global explainability of pediatric dental risk prediction using SHAP.**

This figure combines local and global explainability analyses using SHapley Additive exPlanations (SHAP). The upper panel (waterfall plot) illustrates the contribution of individual socio-demographic and health variables to the predicted dental risk for a representative child. The model baseline prediction ($E[f(X)]$) is progressively shifted toward the final risk estimate ($f(x)$) by feature-specific contributions. Lower income-to-poverty ratio (INDFMPIR) and increasing age (RIDAGEYR) contributed most strongly to elevated predicted risk, followed by race/ethnicity (RIDRETH1) and gender (RIAGENDR), while medical condition history (MCQ010) had minimal influence. The lower panel (SHAP summary plot) displays global feature importance across the study population. Each point represents an individual child, with color indicating feature value (blue = lower values, red = higher values). Age and income-to-poverty ratio emerged as the dominant predictors of dental risk, demonstrating that predictions arise from cumulative socio-demographic effects rather than reliance on a single dominant variable. Together, these plots highlight the transparency and non-deterministic behavior of the model.

## 3.2 Calibration Analysis

Calibration analysis demonstrated that the model produced conservative probability estimates, particularly at higher risk levels. Predicted probabilities tended to underestimate observed risk rather than overestimate it. This conservative behavior is desirable for screening applications, as it minimizes the risk of false alarms and inappropriate clinical escalation[18].

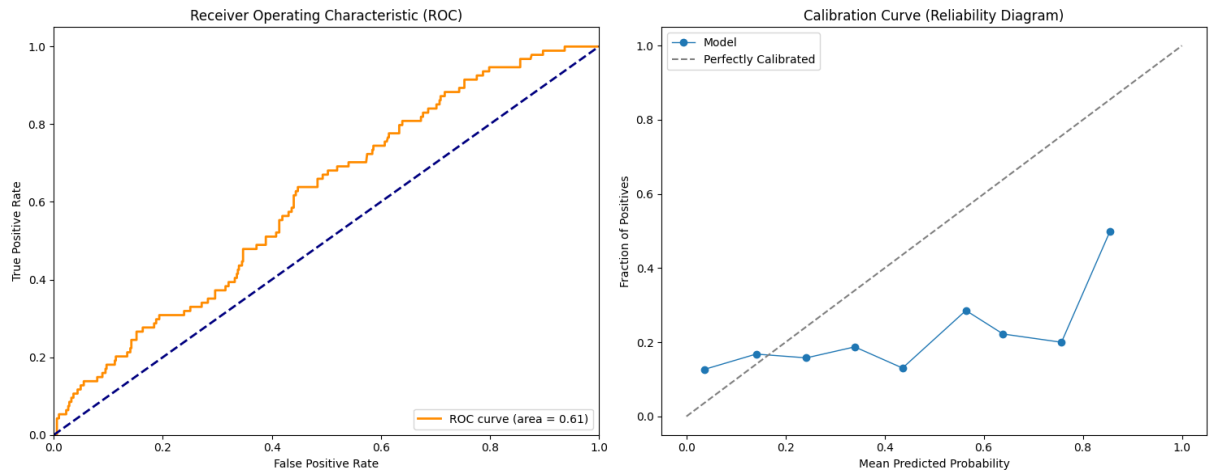

**Figure 2. Receiver operating characteristic (ROC) curve demonstrating discriminative performance of the explainable AI model.**

The receiver operating characteristic (ROC) curve illustrates the ability of the model to distinguish higher-risk from lower-risk pediatric dental outcomes across decision thresholds. The model achieved an area under the curve (AUC) of 0.61, indicating modest discriminative performance. This level of performance reflects the multifactorial and socially mediated nature of pediatric dental risk and is consistent with the intended use of the model as a screening and risk stratification tool rather than a diagnostic classifier.

### 3.3 Global Explainability Findings

Global SHAP analysis identified age and income-to-poverty ratio as the strongest contributors to predicted dental risk. Lower income levels were consistently associated with higher predicted risk. Race/ethnicity and gender showed moderate contributions, while medical history had minimal influence as seen in Figure 3.

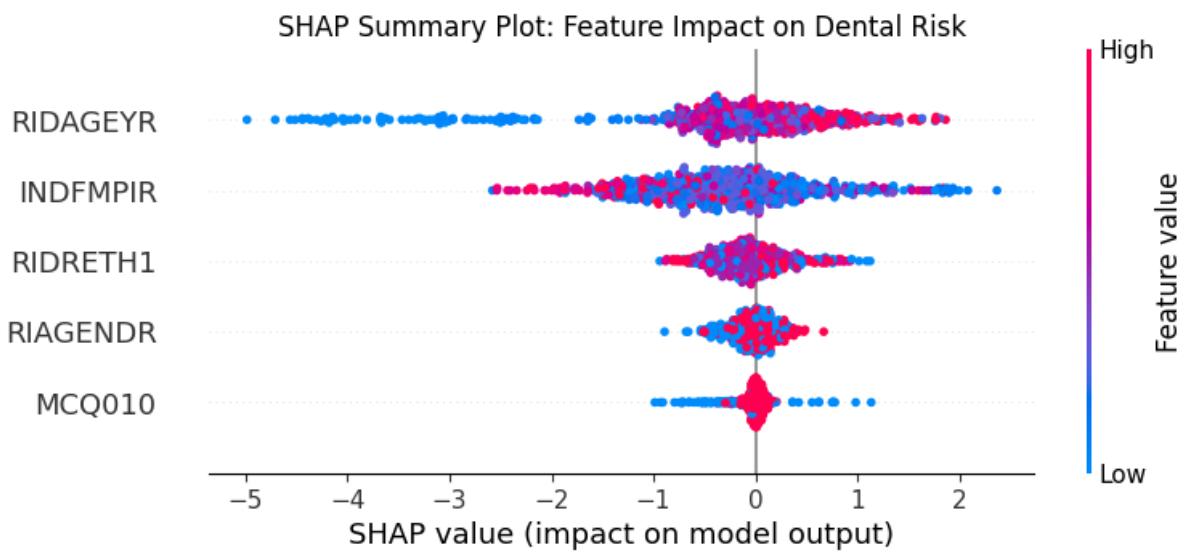

**Figure 3. Calibration curve (reliability diagram) showing agreement between predicted and observed pediatric dental risk.**

The calibration curve compares mean predicted probabilities with observed outcome frequencies across risk bins. The dashed diagonal line represents perfect calibration. The model demonstrates conservative probability estimates, with systematic underestimation at higher predicted risk levels. Such conservative calibration is desirable for screening-oriented applications in pediatric populations, as it minimizes false-positive labeling and supports ethical deployment of risk prediction tools.

These findings align closely with established public health literature emphasizing the dominant role of socioeconomic determinants in pediatric oral health[19-22]

## 3.4 Individual-Level Explainability

Individual SHAP waterfall plots revealed that high-risk predictions resulted from the cumulative influence of multiple socio-demographic factors rather than reliance on a single dominant feature. This pattern underscores the model's non-deterministic behavior and supports its ethical transparency.

## 4. Discussion

This study presents an explainable mind learning artificial intelligence (XAI) framework for pediatric dental risk stratification that prioritizes transparency, ethical alignment, and preventive applicability over maximal predictive accuracy. The findings demonstrate that even modestly performing models can provide meaningful and actionable insights when their reasoning is explicitly revealed and grounded in established public health and epidemiological knowledge.[23],[24]

## 4.1 Interpretation of Predictive Performance

The observed discriminative performance (AUC = 0.61) reflects the inherent complexity of pediatric dental risk rather than a limitation of the modeling approach. Pediatric oral health outcomes are not driven by isolated biological markers but emerge from layered interactions among socioeconomic conditions, caregiver behaviors, access to care, dietary patterns, and environmental exposures.[2],[12] In this context, expecting high discrimination from a small set of socio-demographic variables would be unrealistic and potentially misleading. Importantly, many high-performing AI models reported in the dental literature rely on image-based datasets with tightly curated labels, often collected in controlled clinical environments[4]. While such models may achieve high AUC values, their applicability to population-level screening and prevention is limited. In contrast, the present model operates under real-world constraints using non-clinical variables, aligning more closely with public health deployment scenarios. From a methodological perspective, the modest AUC supports the interpretation of pediatric dental risk as a continuous and probabilistic phenomenon rather than a binary state, reinforcing the framing of the model as a risk stratification and screening tool rather than a diagnostic classifier[3].

## 4.2 Calibration and Conservative Risk Estimation

Calibration analysis revealed that the model systematically underestimates risk at higher predicted probability levels. Although such behavior may be viewed as a limitation in diagnostic contexts, it represents a desirable property for screening-oriented applications. Conservative risk estimation reduces the likelihood of false-positive labeling, which is particularly important in pediatric settings where overestimation of risk may lead to unnecessary anxiety, stigma, or inappropriate intervention[18]. This conservative behavior aligns with ethical principles for AI in healthcare, including harm minimization, proportionality, and explicit acknowledgment of uncertainty. Rather than producing overconfident predictions, the model signals elevated risk cautiously, reinforcing its role as a decision-support system rather than an authoritative decision-maker. Future work may explore post hoc calibration techniques such as isotonic regression or Platt scaling; however, such adjustments should be evaluated carefully to avoid introducing artificial confidence that could undermine ethical deployment.

## 4.3 Explainability as a Primary Contribution

A central contribution of this work lies in its explainability-first design. Global SHAP analysis consistently identified age and income-to-poverty ratio as the dominant contributors to predicted dental risk. This finding closely mirrors decades of epidemiological research demonstrating that socioeconomic status is among the strongest predictors of pediatric oral health outcomes[7,13]. By making these relationships explicit, the model avoids a critical ethical pitfall of black-box AI systems: obscuring structural determinants behind opaque predictions[8]. Rather than presenting risk as an intrinsic property of the child, the model highlights how broader social and economic contexts shape oral health outcomes. Individual-level SHAP analyses further demonstrated that high-risk predictions arose from the cumulative influence of multiple socio-demographic factors rather than reliance on a single variable. This non-deterministic behavior is particularly important in pediatric applications, as it reduces the risk of stigmatization or oversimplified interpretations based on isolated attributes such as race or gender[6].

## 4.4 Implications for Health Equity and Bias

The prominence of race/ethnicity and income-related variables in model explanations warrants careful interpretation. These variables should not be understood as biological determinants of dental risk, but rather as proxies for structural inequities, including differential access to care, environmental exposures, and historical marginalization. By explicitly surfacing these factors, the model enables critical reflection rather than silent bias propagation. In contrast to black-box systems that may encode disparities without disclosure, the explainable framework allows clinicians, policymakers, and researchers to interrogate whether—and how—such variables should inform intervention strategies. This transparency supports ethical deployment by shifting the focus from individual blame to systemic responsibility. For example, high predicted risk associated with low income may justify targeted preventive programs, expanded access to fluoridation, school-based sealant initiatives, or community-level interventions rather than individual-level punitive measures[17].

## 4.5 Clinical and Public Health Relevance

Although the model is not intended for clinical diagnosis, it has several potential applications in pediatric dentistry and public health. These include population-level screening to identify groups that may benefit from intensified preventive care, prioritization of limited public health resources, and evidence-informed policymaking aimed at reducing oral health disparities.

Importantly, the model's explainability facilitates interdisciplinary engagement. Pediatric dentists, public health professionals, caregivers, and policymakers can interpret model outputs in context, understanding not only *what* risk is predicted but *why*—a prerequisite for trust and responsible adoption in healthcare settings[11].

## 4.6 Comparison with Existing AI Approaches in Dentistry

Most existing AI applications in dentistry emphasize detection tasks, such as identifying caries on radiographs or classifying oral pathology images. While valuable, these approaches typically operate downstream of disease development and offer limited insight into upstream risk mechanisms. In contrast, the present study focuses on risk emergence rather than disease detection**,** aligning more closely with preventive dentistry and public health objectives. By embedding explainability at the core of the modeling process, the framework addresses growing concerns about transparency, accountability, and bias in medical AI[5].

## 4.7 Limitations and Future Directions

Several limitations should be acknowledged. First, the cross-sectional nature of the dataset precludes causal inference and limits modeling of temporal risk trajectories. Longitudinal data would allow for more nuanced analysis of how socio-demographic risk accumulates over time. Second, the absence of clinical dental examination findings and behavioral variables constrains predictive performance. Future studies could integrate clinical indicators, dietary information, or oral hygiene behaviors, provided that explainability is preserved. Finally, while SHAP provides a robust framework for model interpretation, it reflects associations learned by the model rather than causal relationships. Caution is therefore required to avoid overinterpreting feature importance as evidence of causation[20,22].

## 4.8 Broader Implications for Responsible AI in Pediatrics

More broadly, this work illustrates a paradigm for responsible AI development in pediatric healthcare. Rather than pursuing maximal accuracy at the expense of transparency, the study demonstrates how modestly performing models can still deliver substantial value when their reasoning is explicit, cautious, and ethically framed. This approach aligns with emerging international guidelines for trustworthy AI, which emphasize interpretability, fairness, and accountability as core requirements for healthcare applications. In pediatric settings—where vulnerability, long-term consequences, and equity are paramount—such principles are especially critical.

## 5. Conclusion

This work presents a transparent, explainable AI framework for pediatric dental risk stratification that prioritizes ethical deployment and preventive applicability. By making socio-demographic drivers of risk explicit, the model supports equitable, responsible use of AI in pediatric dental public health. This study is limited by its cross-sectional design and the absence of clinical dental examination data, behavioral variables, and longitudinal follow-up. Future work should explore longitudinal modeling and multimodal integration while preserving explainability.